\documentclass[journal,]{IEEEtran}

\usepackage{enumerate}
\usepackage{cite}
\usepackage{tikz}
\usepackage{xcolor}
\graphicspath{{figures/}}
\usepackage{algorithm }
\usepackage{algorithmic}

\usepackage{makecell}
\usepackage{bm}
\usepackage{amsmath,amsfonts,amssymb,amsthm}
\usepackage{subfigure}
\usepackage{threeparttable}
\usepackage{booktabs}
\usepackage{bm}
\usepackage{mathrsfs}
\usepackage{flushend}

\usepackage{url}
\usepackage{hyperref}

\usepackage{footnote}
\makesavenoteenv{tabular}

\begin{document}

\begin{table*}
\centering
\begin{tabular}{p{18cm} }
\\ This paper has been accepted for publication in IEEE Internet of Things Journal\\  \\ 

\\ DOI: 10.1109/JIOT.2019.2957778\\  \\ 

\\ IEEE Explore \url{https://ieeexplore.ieee.org/abstract/document/8924617} \\ \\ 

\\ \\Please cite the paper as: \\

\\ Y. Li, X. Hu, Y. Zhuang, Z. Gao, P. Zhang and N. El-Sheimy, "Deep Reinforcement Learning (DRL): Another Perspective for Unsupervised Wireless Localization," in IEEE Internet of Things Journal, doi: 10.1109/JIOT.2019.2957778, 2019. \\

\\ \\ bibtex:  \\

\\ @article\{LiY-DRL-Loc, \\
  author=\{Y. \{Li\} and X. \{Hu\} and Y. \{Zhuang\} and Z. \{Gao\} and P. \{Zhang\} and N. \{El-Sheimy\}\},\\
  journal=\{IEEE Internet of Things Journal\}, \\
 title=\{Deep Reinforcement Learning (DRL): Another Perspective for Unsupervised Wireless Localization\},\\ 
  year=\{2019\},\\
  volume=\{\},\\
  number=\{\},\\
  pages=\{1-1\},\\
  doi=\{10.1109/JIOT.2019.2957778\},\\
  ISSN=\{2327-4662\},\\
  month=\{\}\,\\
  \}\\

 \end{tabular}
\end{table*}

\clearpage

\title{ Deep Reinforcement Learning (DRL): Another Perspective for Unsupervised Wireless Localization  \thanks{
}}
\author{You Li,~\IEEEmembership{Member,~IEEE},
  Xin Hu,~\IEEEmembership{Senior Member,~IEEE},
  Yuan Zhuang,~\IEEEmembership{Member,~IEEE}, \\
  Zhouzheng Gao,
  Peng Zhang,  
  and Naser El-Sheimy
   \thanks{Y. Li and N. El-Sheimy are with Department of Geomatics Engineering, University of Calgary (liyou331@gmail.com; elsheimy@ucalgary.ca). X. Hu is with the School of Electronic Engineering, Beijing University of Posts and Telecommunications (huxin2016@bupt.edu.cn). Y. Zhuang and P. Zhang are with the State Key Laboratory of Surveying, Mapping and Remote Sensing, Wuhan University (zhy.0908@gmail.com; fenix@whu.edu.cn). Z. Gao is with the Department of Land Sciences, China University of Geosciences (Beijing) (zhouzhenggao@126.com). Y. Li and X. Hu contributed equally to this work. Corresponding author: X. Hu.
   
   This paper is partly supported by the Natural Sciences and Engineering Research Council of Canada (NSERC) CREATE Grants, NSERC Discovery Grants, NSERC Strategic Partnership Grants, the Alberta Innovates Technology Future (AITF) Grants, the Canada Research Chair (CRC) Grants, and the National Natural Science Foundation of China (NSFC) Grants (No. 41804027, 61771135, and 61873163).}
}

\markboth{IEEE Internet of Things Journal}%
         {Shell \MakeLowercase{\textit{et al.}}: Bare Demo of IEEEtran.cls for Journals}

         \maketitle

         \begin{abstract}
Location is key to spatialize internet-of-things (IoT) data. However, it is challenging to use low-cost IoT devices for robust unsupervised localization (i.e., localization without training data that have known location labels). Thus, this paper proposes a deep reinforcement learning (DRL) based unsupervised wireless-localization method. The main contributions are as follows. (1) This paper proposes an approach to model a continuous wireless-localization process as a Markov decision process (MDP) and process it within a DRL framework. (2) To alleviate the challenge of obtaining rewards when using unlabeled data (e.g., daily-life crowdsourced data), this paper presents a reward-setting mechanism, which extracts robust landmark data from unlabeled wireless received signal strengths (RSS). (3) To ease requirements for model re-training when using DRL for localization, this paper uses RSS measurements together with agent location to construct DRL inputs. The proposed method was tested by using field testing data from multiple Bluetooth 5 smart ear tags in a pasture. Meanwhile, the experimental verification process reflected the advantages and challenges for using DRL in wireless localization.           
\end{abstract}
		 

         \IEEEpeerreviewmaketitle

\section{Introduction}  
\label{sec-intro}
         \IEEEPARstart{T}{he} internet-of-things (IoT) technology has started to empower the future of numerous fields \cite{Zanella-IoT-2014}. To spatialize IoT data, the time and location information of IoT devices are essential. Thus, localization is both an important application scenario and a development direction for IoT. 
        
        IoT localization methods have been widely researched. There are technologies including wireless \cite{Jiang-Pei-2016}, motion, and environmental \cite{Zhou-Liu-2019} sensor based localization, as well as their integration \cite{LiY-thesis}. During the recent decade, the development in IoT technologies and the emergence of geo-spatial big data have made it possible to implement wide-area mass-market localization by using crowdsourced data. However, the performance of such mass-market localization techniques may be degraded by various factors, such as the complexity of localization environment \cite{Zhou-Chen-2017}, the existence of device diversity \cite{LiY-SensJ-2019}, and the uncertainty in crowdsourced data \cite{LiY-JIOT-2019}. Thus, it is still an open challenge to use low-cost IoT devices for robust localization. 
        
        \subsection{Deep-Learning-Based Localization}
		\label{sec-ai-loc}
		The development of deep-learning (DL) techniques have led to the emergence of new localization methods. Examples of such methods include localization using deep neural network (DNN) \cite{Zhang-Liu-2016}, Gaussian processes (GP) \cite{Ferris-2006}, random forests \cite{GuoX-2018}, hidden Markov model (HMM) \cite{SunS-2019}, support vector machine (SVM) \cite{Timoteo-2016}, and fuzzy logic \cite{Orujov-2018}. These DL techniques have also been used in other localization-related aspects. For example, DNN has been used for localization parameter tuning \cite{Chiang-NN}, activity recognition \cite{ZhangX2018}, and localization uncertainty prediction \cite{LiY-NN}. 
		
		
		DL algorithms have shown great potentials in enhancing localization, especially in complex scenarios that are difficult to model, have parameters that are difficult to set, and have nonlinear and correlated measurements. However, most of the existing DL-based localization methods are supervised methods. That is, these methods require training data that have known location labels. The acquisition of location labels is commonly time-consuming and label-costly \cite{Bolliger2008}. Meanwhile, the accuracy of location labels is degraded by factors such as device diversity \cite{LiY-SensJ-2019}, device motion and orientation \cite{Chen-Pei-2010}, and database outage \cite{Solin-2018}. Thus, unsupervised localization methods are needed to reduce reliance on location labels.
		
		\subsection{Unsupervised Localization}
		\label{unsupervised-loc}
		To realize unsupervised localization, researchers have proposed various methods, such as simultaneous localization and mapping (SLAM) \cite{Bruno-L--2011} and crowdsourcing \cite{ZhouB-ITS-2015}. In such methods, the uncertainty in reference point (RP) location labels will directly lead to errors in the generated localization databases. Inertial-sensor-based dead-reckoning (DR) can provide autonomous indoor/outdoor localization solutions \cite{YL-floor-2018}. However, it is challenging to obtain long-term accurate DR solutions with low-cost sensors due to the requirement for heading and position initialization, the misalignment angles between human body and device, and the existence of sensor errors \cite{LiY-sensj-calibration}. Thus, constraints are needed to constrain DR drifts. Vehicle-motion constraints (e.g., zero velocity updates, zero angular rate updates, and non-holonomic constraints) \cite{LiY-thesis} are typically used to correct for DR errors. However, these motion constraints are relative constraints, which can only mitigate the accumulation of DR errors, instead of eliminate them. DR solutions always drift when external updates, such as loop closures and global navigation satellite systems (GNSS) positions, are not available.
		
		In many crowdsourcing applications, it is difficult to assure the reliability of RP locations in the database due to the limitation of physical environment. For example, there may be insufficient observations for wireless localization. If this is the case, it is important to evaluate the quality of localization data, so as to select the robust ones. From the big-data perspective, a small proportion of crowdsourced data, if robust, is enough for database training. The research in \cite{ZhangP2018} presents a general framework for assessing sensor data quality. The evaluation framework involves the impact of indoor localization time, user motion, and sensor biases. Furthermore, the research \cite{LiY-JIOT-2019} enhances this framework and introduces stricter quality-assessment criteria. 
		
		Compared to these works, this research is carried out from another perspective. The extensively-concerned deep reinforcement learning (DRL) technique is applied. DRL has been proven to have the following advantages \cite{HuX-COMM-2018} in other areas: (1) it can be used for unsupervised learning through an action-reward mechanism and (2) it can provide not only the estimated solution at the current moment, but also the long-term reward. Thus, it may bring benefits into the localization field. 

		\subsection{DRL in Navigation and Localization}
		\label{sec-drl-nav-loc}
		DRL, which is the core artificial intelligence (AI) algorithm for the AlphaGo, has attracted intensive attention. DRL can be regarded as a combination of DL and reinforcement learning (RL). The former provides learning mechanisms, while the later sets goals for learning. In general, DRL involves agents that observe states and act in order to collect long-term rewards \cite{Hu-Jia-DRL-2019}. The DRL algorithm has experienced stages such as the deep Q-network (DQN), asynchronous advantage actor-critic (A3C), and unsupervised reinforcement and auxiliary learning (UNREAL) \cite{Jaderberg-Mnih-2017}. The research in \cite{HuX-COMM-2018} points out three components for a DRL solution: basis/core (e.g., the definition of states, actions, and reward function), basic units (e.g., the Q-network, action selection, replay memory, and target network), and state reformulation (i.e., the method for state-awareness data processing). 
	\begin{table}
\centering
\begin{tabular}{p{1.6cm} p{5.8cm} }
 \hline
\textbf{Abbreviation} & \textbf{Definition}  \\ \hline
AI & Artificial Intelligence  \\ 
A3C &  Asynchronous Advantage Actor-Critic  \\ 
BT5 & Bluetooth 5  \\ 
CDF & Cumulative Distribution Function  \\ 
DL & Deep Learning \\ 
DNN & Deep Neural Network \\ 
DQN & Deep Q-Network \\ 
DR &  Dead-Reckoning  \\ 
DRL & Deep Reinforcement Learning \\ 
GNSS &  Global Navigation Satellite Systems \\ 
GP & Gaussian Processes \\ 
GW  & Gateway \\ 
HMM & Hidden Markov Model \\ 
ID & IDentification \\ 
IoT &  Internet of Things \\ 
LF & Localization Feature  \\ 
MDP & Markov Decision Process \\ 
 NLoS & Non-Line-of-Sight \\ 
 RL & Reinforcement Learning \\ 
 RMS  & Root Mean Square \\ 
 RP & Reference Point \\ 
RSS & Received Signal Strength \\ 
SLAM &  Simultaneous Localization And Mapping \\ 
SVM &Support Vector Machine \\ 
UNREAL & UNsupervised REinforcement and Auxiliary Learning \\ \hline
 \end{tabular}
\caption{ List of abbreviations   }
\label{tab:abbre}
\end{table}
		
		Navigation is an important application scenario for DRL. The early-stage DRL algorithms are used for learning in vedio games. Such gaming applications require navigation actions in a virtual world. Another classic scenario for DRL research is maze navigation \cite{Mirowski-Pascanu-2017}. The research in \cite{Dhiman-Banerjee-2019} provides deep investigation on the performance of DRL-based maze navigation. Furthermore, the research on DRL-based navigation has been extended from visual to real world. Researchers have utilized DRL for navigation by using data from various types of sensors, such as camera \cite{Zhang-Springenberg-2019}, lidar \cite{Tai-Paolo-2017}, 360-degree camera \cite{Bruce-Sunderhauf-2017}, Google street view \cite{Mirowski-Grimes-2018}, wireless sensors \cite{Mohammadi-Fuqaha-2018}, and magnetic sensors \cite{BejarE-2018}. Meanwhile, other data or techniques, such as topological maps \cite{Kato-Kato-2017}, particles \cite{Zhao-Braun-2017}, cooperative agents \cite{PengB-2019}, and social interactions \cite{Chen-Everett-2018} have been involved. The latest directions for DRL-based navigation include mapless navigation \cite{Tai-Paolo-2017}\cite{Bruce-Sunderhauf-2017}\cite{Zhelo-Zhang-2018}, navigation in new \cite{Zhelo-Zhang-2018} and complex \cite{Mirowski-Pascanu-2017} environments, and navigation with varying targets \cite{Zhu-Mottaghi-2018}.

		\subsection{Problem Statement and Main Contributions}
        \label{sec-contribution}
        The DRL-based approaches have been verified to be effective in navigation. However, most of these methods are not suitable for localization. Although navigation and localization are not separated in many applications, they have different principles. Navigation and localization may use the same input (e.g., signals from wireless, image, and environmental sensors) but have different outputs. Navigation is the problem of finding the optimal path between the agent (i.e., a IoT end-device) and a target place; thus, its output is the moving action. In contrast, the output for localization is the agent location. The challenges for using DRL for localization include

        \begin{itemize}
        \item In a navigation application, the agent chooses an action from the DRL engine and move. The action directly changes the state (i.e., the agent location). Thus, a navigation process can be modeled as a Markov decision process (MDP) and thus can be processed by DRL. However, localization is closer to a DL problem, instead of DRL. 
        \item The existing DRL-based localization methods (e.g., \cite{Mohammadi-Fuqaha-2018}) require target points, which are necessary for setting rewards. To obtain such target points, supervised or semi-supervised data are needed. 
        \item The existing target-dependent navigation and localization methods suffer from another issue; that is, the trained model is related with the target. When the target changes, re-training may be needed. This phenomenon also limits the use of DRL in localization. 
        \item Most of the existing works are based on vision data. Thus, it is necessary to investigate the use of DRL in wireless positioning, which is the most widely used technology in IoT localization.
        \end{itemize}

	Therefore, the main contributions of this paper can be stated as follows.
	\begin{itemize}
	\item It is difficult to use DRL in traditional snapshot localization models because these models do not meet the MDP definition. Thus, this paper proposes a method to model a continuous wireless localization process as an MDP and process it within a DRL framework.
	\item It is challenging to obtain DRL rewards when using only unsupervised data (e.g., daily-life crowdsourced data). To alleviate this issue, this paper presents a reward-setting mechanism for unsupervised wireless localization. Robust landmark data are extracted automatically from unlabeled wireless received signal strengths (RSS).
	\item To ease requirements for model re-training when using DRL for localization, this paper uses RSS measurements together with the agent location to construct the input for DRL. Thus, it is not necessary to re-train DRL when the target changes. 
	\end{itemize}

\section{DRL-based Unsupervised Wireless Localization}
\label{sec-methodology}
 This section describes the methodology of DRL-based unsupervised wireless localization. Specifically, this section is comprised of the problem description, the construction of MDP model for wireless localization, the details of the DQN algorithm, and the mechanism for reward setting.
        
\subsection{Problem Description}    
\label{sec-problem-description}
The purpose for localization is to determine the agent position in a spatial coordinate system. The agent position can also be represented by gridding the space and determining the identification (ID) of the grid that the agent is located in. To determine the agent location, surrounding localization signals (LFs) such as RSS are measured. The fingerprinting method is commonly used for localization through two steps, training and prediction. At the training step, [location, LF] fingerprints at multiple RPs are used to generate a database. At the prediction step, the likelihood value between the real-time measured LF vector and the reference LF vector at each RP in the database is computed. The RPs with the LFs that are closest to the measured one are selected to compute the agent location \cite{Haeberlen2004}. From this perspective, localization is a DL problem, which inputs LF measurements and outputs the RP ID.

The fingerprinting method provides a snapshot localization solution. Its advantage is that a location output can be obtained once a real-time LF measurement is inputted. For dynamic localization applications, a common approach is to further input the snapshot localization solution into a localization filter (e.g., an extend Kalman filter or particle filter) to generate a more robust solution by fusing the previous location solutions. In the filter, the snapshot localization solutions are position updates, while sensor-based DR data or pseudo motion constraints (e.g., the constant-velocity assumption) are used to construct the system motion model \cite{LiY-thesis}.

\begin{figure}
\centering
\includegraphics[width=0.43 \textwidth]{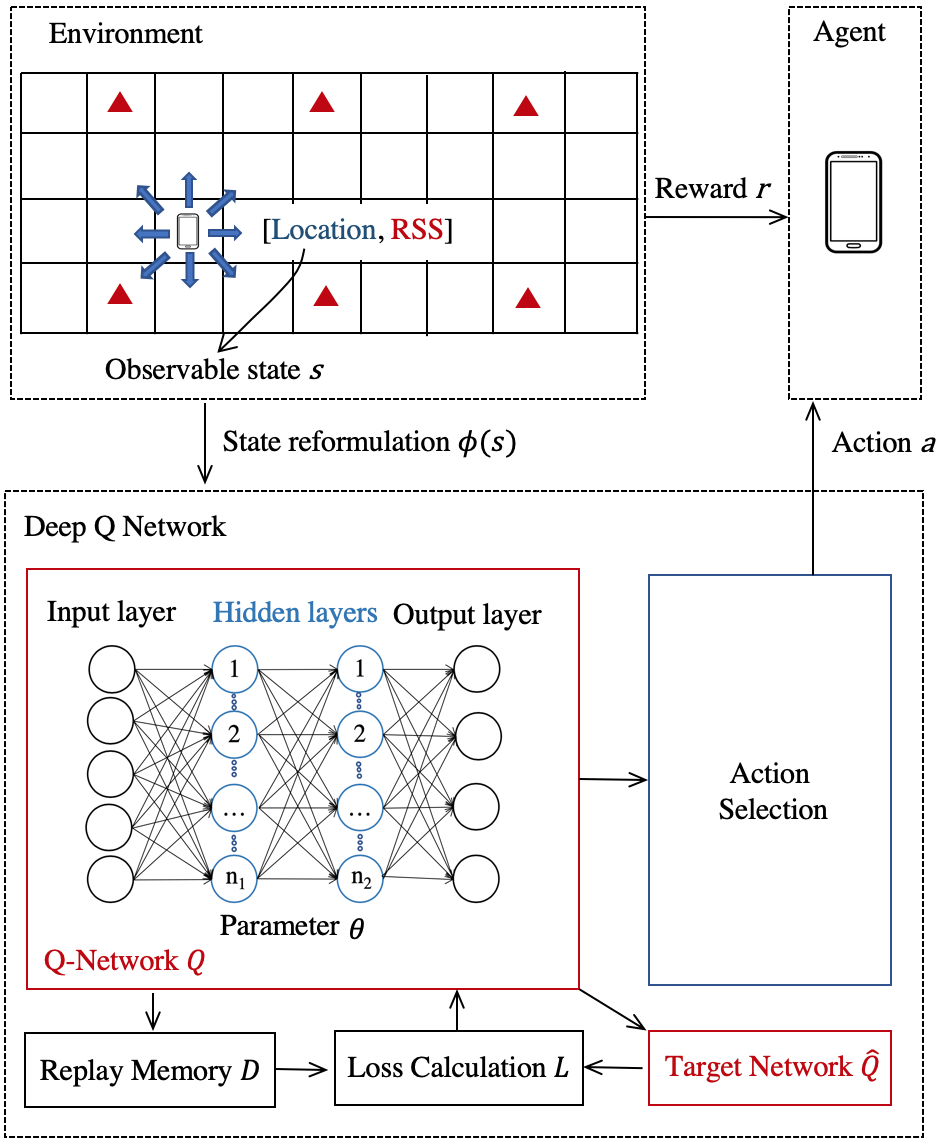}
\caption{Architecture for DRL-based Wireless Localization. Red triangles indicate wireless gateway locations}
\label{fig:dqn}
\end{figure}

This research changes the wireless localization process by introducing the previous location solutions. Accordingly, wireless localization becomes a continuous localization problem. The localization task at time $t$ is the process that inputs the agent location at time $t-1$ plus the LF measurement at time $t$, and outputs the agent location at time $t$. After the localization computation at time $t$, the agent may keep static or move towards one of the eight directions in Figure \ref{fig:dqn}. Afterwards, the localization computation at time $t+1$ starts. In this case, the localization computation at each time step only depends on the location from the previous time step and the LF measurement at this time step; meanwhile, the action at each step directly changes the location state. Thus, this process can be modeled as an MDP.
        
\subsection{MDP Model}
\label{sec-mdp}
An MDP is a discrete-time stochastic control process. Its current state is only related with the latest previous state, instead of earlier ones. In contrast to the Markov chain and HMM, the MDP has involved actions, which directly influent states. An MDP is comprised of four components: states $ s_t \in S$, actions $ a_t \in A$, a reward function $r \in \mathbb{R}$, and transition probabilities $ p (s_{t+1}|s_t, a_t) $ of moving from $s_t$ to $s_{t+1}$ given $a_t$, where $s_t$ and $a_t$ are the state and action at time step $t$, respectively. The goal for an MDP is to determine the policy that maximizes the expected accumulated rewards $ R_t = \sum_{i=1}^{ \infty } { (\gamma^i r_{t+i} ) } $, where $r_{t+i}$ is the immediate reward at time step $t+i$ and  $\gamma \in [0,1]$ is the discount factor \cite{LiuS-2018}. Figure \ref{fig:dqn} demonstrates a schematic diagram for DRL-based wireless localization. The state, action, and reward definition have been shown. The details of the components in the figure are described in this subsection and Subsection \ref{sec-meth-dqn}. To design an MDP for wireless localization, the following three components are defined.

States: the state $s_t$ represents an abstraction of the environment in which the agent makes action decisions at time $t$. The state consists of the agent location and the RSS measurement.

Actions: the agent makes decisions to take actions based on the state $s_t$. In this research, the action space consists of nice actions, including staying at the same grid and moving toward north, south, west, east, northwest, northeast, southwest, and southeast for a grid.

Reward function: a positive reward will be given when the agent has made a correct action. Theoretically, the geographical distance between the agent and target point can be used for setting rewards \cite{Mohammadi-Fuqaha-2018}. This mechanism is effective when using supervised or semi-supervised data; however, it cannot be used to process unsupervised data. To alleviate this issue, a reward-setting mechanism is presented. The principle of this mechanism is to extract landmark points that have robust location labels and RSS features. When the agent has moved to a landmark point and the measured RSS has the similar feature to the known RSS feature at this landmark point, a positive reward is set. 

A challenge for this mechanism is that it is difficult to know either the location or the RSS feature at a landmark point in advance. To alleviate this issue, the locations of wireless gateways (GWs, also known as access points or anchors) and the near-field condition are introduced. Specifically, the near-field condition is activated when it is detected that the agent has moved to a location that is close enough to a GW. Then, the distance between the predicted agent location and the location of this GW is used to set the reward. The method for detecting the near-field condition is described as follows. 

One of the most widely-used approaches for detecting the near-field condition is RSS ranging. The wireless signal path-loss model is widely used to convert an RSS to an agent-GW distance $d$ by 
\begin{equation}
d = 10 ^ { \frac{RSS - b}{-10 n} }
\end{equation}
where $n$ and $b$ are the path-loss-model parameters. Although such parameters can be trained in advance \cite{Zhuang-Y-EL-2015}, there are various factors (e.g., device diversity and orientation diversity \cite{LiY-SensJ-2019}) that may cause variations in these parameters. This phenomenon leads to the degradation in RSS-based ranging and localization accuracy. Thus, it is challenging to detect the near-field condition through RSS ranging. 

To alleviate this issue, the following phenomenon is used: environmental and motion factors commonly weaken an RSS measurement, instead of strengthen them. Therefore, a weak RSS measurement does not ensure a long distance; in contrast, a strong RSS can indicate a short distance. Accordingly, the near-field condition can be identified as: when the measured RSS from a GW is stronger than a threshold $\beta_{R}$, the agent should be located near this GW. Then, the reward $r_t$ can be set as 
\begin{equation}
r_t =\left\{
\begin{aligned}
 & \frac{1}{d_{t,i}},~ \mathrm{if} ~ RSS_i > \beta_{R} ~ \& ~ d_{t,i}  \leq \beta_{d} \\
 & -d_{t,i},~ \mathrm{if} ~ RSS_i > \beta_{R} ~ \& ~ d_{t,i} > \beta_{d} \\
 & 0,~\mathrm{otherwise}
\end{aligned}
\right.
\end{equation}
         where $d_{t,i}$ is the distance between the location of the agent at time $t$ and that of the $i$-th GW; $\beta_{d}$ is the threshold for the distance between the predicted agent location and the location of the selected GW. The case $d_{t,i} > \beta_{d}$ indicates that the agent is wrongly located to a point that is far from the landmark point; thus, a negative reward is set.
         
         The states, actions, and reward function are further used for training of the DQN, which is described in the next subsection.
         
\subsection{Deep Q-Network for Wireless Localization}
\label{sec-meth-dqn}         
A core of DQN is Q-learning. The principle of Q-learning is to determine the $Q$ function 
\begin{equation}
Q:~\phi(s_t) \rightarrow Q\left(   \phi(s_t), a_t; \theta    \right)
\end{equation}
  which can be used to compute the expected accumulated rewards for taking an action $a_t$ when there is a given input $\phi (s_t) $, where $\theta$ is the action-value function that maps the input to output decisions; $\phi (s_t) $ is the state reformulation. Once the $Q$ function is obtained, it becomes possible to construct a policy $\pi (s) $ that maximizes the rewards by 
\begin{equation}
 \pi (s) = \underset{a} {\mathrm{argmax}} ~Q(s,a)
\end{equation}

For applications (e.g., navigation in a simple grid maze) that have a simple state, matrix-based equations may be used to compute $Q$. For the task in this research, it is challenging to model the Q-learning process. Thus, a DNN is used to resemble Q. The DQN architecture in \cite{HuX-COMM-2018} is used. The DQN algorithm is shown in Table \ref{tab:dqn-algorithm}.

\begin{table}
\centering
\begin{tabular}{@{}p{0.01cm}p{0.01cm}p{8.0cm}@{}}
  \hline
  \multicolumn{3}{l}{\textbf{Algorithm 1:} The DQN algorithm (modified on \cite{HuX-COMM-2018})} \\
  \hline
1.    &&      Initialize replay memory $D$, Q-network $Q$, and target network $\hat{Q}$; \\ 
2.    &&      For time step t in 1 to T: \\
3.    &&     ~~~~ Observe observable state $s_t$ and set reward $r_t$; \\
4.    &&     ~~~~ Generate state reformulation $\phi (s_t) $;            \\
5.    &&     ~~~~ Stack experience tuple $\left( \phi (s_{t-1}), a_{t-1}, r_t,\phi (s_{t})  \right) $ into $D$;            \\
6.    &&     ~~~~ Compute available action set $A (s_{t}) $;            \\
7.    &&     ~~~~ With exploration probability $\epsilon$:              \\
8.    &&     ~~~~  ~~~~  Select a random action $a_t$ in $ A (s_{t}) $;            \\
9.    &&     ~~~~ Otherwise:                      \\
10.    &&     ~~~~  ~~~~  Select $a_t=  \underset{a \in A (s_{t})} {\mathrm{argmax}} ~ Q (\phi (s_t), a;\theta ) $;            \\
11.    &&     ~~~~ Move agent by action $a_t$;            \\
12.   &&     ~~~~   Sample a minibatch of $(\phi (s_j), a_j, r_{j+1}, \phi (s_{j+1}) )$ from $D$;          \\
13.   &&     ~~~~  Compute target value $y_j$ through \eqref{eq-target-value};           \\
14.   &&     ~~~~  Compute loss through \eqref{eq-loss};           \\
15.   &&     ~~~~  Train Q-network through SGD;           \\
16.   &&     ~~~~  Decrease exploration probability $\epsilon$;           \\
17.   &&     ~~~~  if t modulo G == 0:          \\
18.   &&     ~~~~  ~~~~ Update target network $\hat{Q}$ with $\theta ^- = \theta$;           \\
19.   &&     End For loop             \\
  \hline
\end{tabular}
\caption{DQN Algorithm}
\label{tab:dqn-algorithm}
\end{table}
         
     During each mapping from the input to the output decision, the Q-network generates a result that consists the current state $\phi (s_j) $, the current action $a_j$, the instant reward $r_{j+1}$, and the next state $\phi (s_{j+1}) $. Such a result is then stored into the replay memory $D$. The target network $\hat{Q}$ with parameter $\theta^{-}$ is copied from the Q-network in every $G$ steps. At each step, a minibatch is sampled randomly from the replay memory $D$ and combined with the target network $\hat{Q}$ to compute the loss and train the Q-network. 
     
     The replay memory $D$, which has a capacity of $N_{ep}$ is created at the initialization step. Afterwards, the newly-generated experience tuple $\left( \phi (s_{t}), a_{t}, r_{t+1},\phi (s_{t+1})  \right) $ is stacked into $D$. The Q-network is trained when the length of the stored experience tuples reaches the number $N_{st}$. For training, a minibatch that has a length of $N_{mb}$ is sampled randomly from $D$. Meanwhile, for each time step in training, the epsilon-greedy policy is used to select actions. The epsilon-greedy policy also balances the reward maximization based on the already-known knowledge (i.e., the exploitation) and the new knowledge that is obtained by trying new actions (i.e., the exploration). The exploration rate $\epsilon$ is decreased linearly from the initial value $\epsilon_i$ to final value $\epsilon_f$ during training.
For each experience tuple within the sampled minibatch, the target network $\hat{Q} (\phi (s), a;\theta^{-}) $ is used to compute the loss $L (\theta) $ as 
\begin{equation}
\label{eq-loss}
L (\theta) = E[(y_j-Q( \phi(s),a;\theta_j )    )^2]
\end{equation}
    where the sign $E[\cdot]$ represents the computation of expectation value; $y_j$ is the target value, which can be calculated as
     \begin{equation}
\label{eq-target-value}
y_j = r_{j+1} + \gamma~\underset{a} {\mathrm{max}} ~ \hat{Q} ( \phi(s_{j+1},a; \theta^-)  )
\end{equation}
        
        Once the loss value is computed, the stochastic gradient descent (SGD) method \cite{SGD} is applied to train the Q-network. During the training process, the batch-normalization approach \cite{Ioffe-Szegedy-2015} is applied to accelerate training.
        
\section{Experimental Verification}
\label{sec-test}

\subsection{Test Description}
\label{sec-test-des}
\begin{figure}
\centering
\includegraphics[width=0.45 \textwidth]{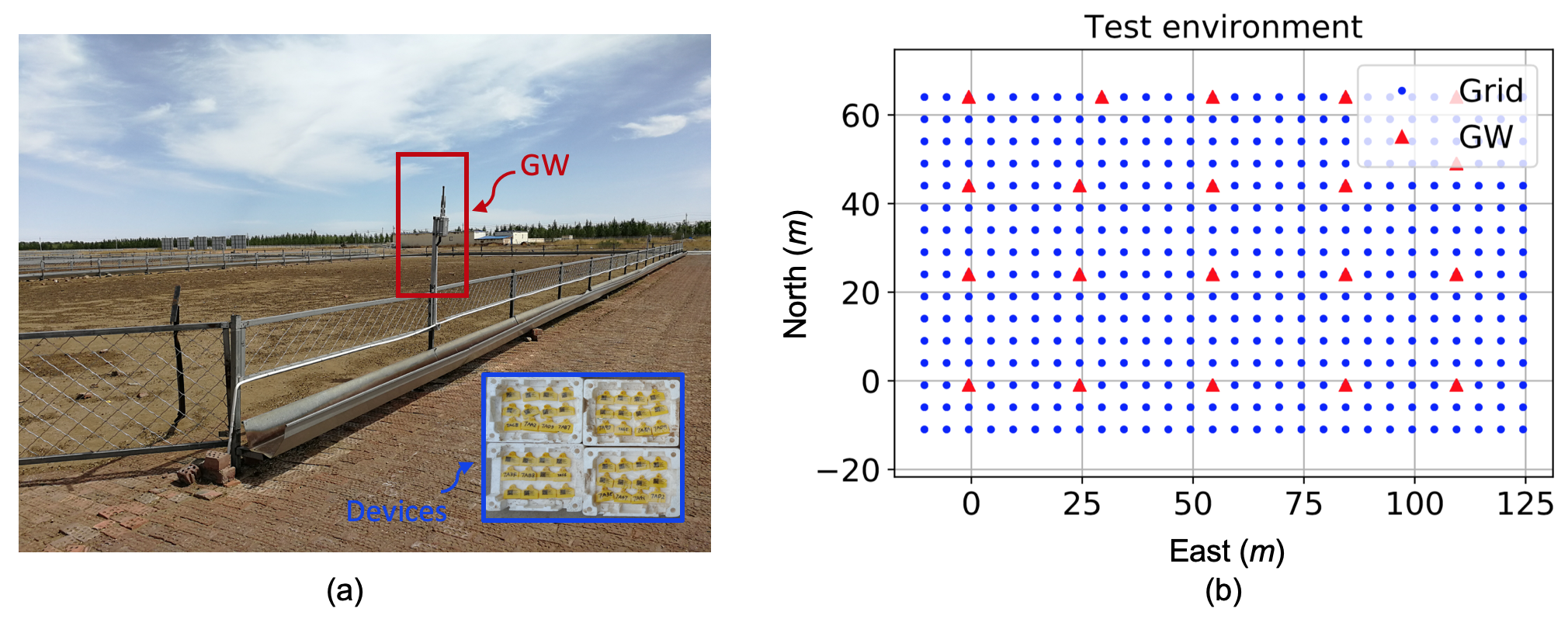}
\caption{Test field and devices (a) and locations of grids and GWs (b)}
\label{fig:test-environment}
\end{figure}

Field tests were carried out in a smart pasture at Inner Mongolia, China. The test area was an open field that had a size of 120 m by 70 m. The test scenario was similar to that in \cite{LiY-SensJ-2019}. Figure \ref{fig:test-environment} (a) demonstrates the test environment and devices. Totally 48 Bluetooth 5 (BT5) based devices (i.e., smart ear tags) were utilized as transmitters, while 20 GWs were used as receivers. Both the devices and GWs were equipped with the Texas Instruments CC2640R2F BT5 chips \cite{TI-CC2640R2F}. Each device was equipped with a microstrip patch antenna with a gain of 0 dBi, while each GW was equipped with a vertical-polarized omni-directional antenna with a gain of 5 dBi. The GWs were deployed evenly over the space by 4 rows and 5 columns. The distances between adjacent GWs were approximately 30 m in the east and 24 m in the north.

\begin{figure}
\centering
\includegraphics[width=0.49 \textwidth]{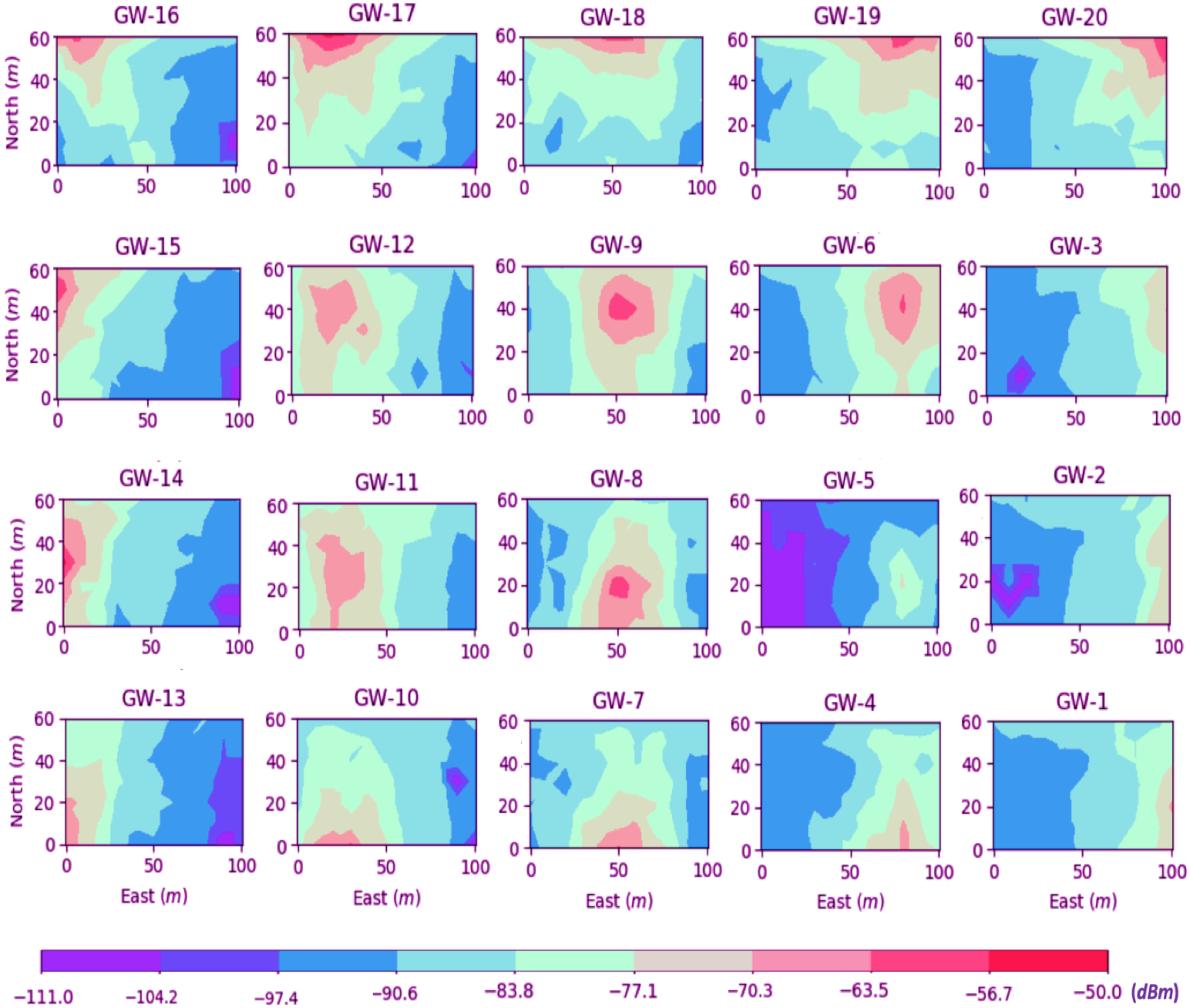}
\caption{GW IDs and distribution heatmaps of RSS measurements}
\label{fig:rss-heatmap}
\end{figure}

The devices were placed at 950 static points on the ground, each for 5 minutes. The data rate for RSS measurements was 0.17 Hz. The data collection process was conducted through a supervised procedure. That is, each data sample had a reference location label. The location labels were only used for localization performance evaluation, instead of localization computation. For this research, the location information in collected data was evenly gridded into 448 grids (i.e., in 16 rows and 28 columns, each grid had a size of 5 m by 5 m), that is, each location data was replaced by that at the nearest grid. Figure \ref{fig:test-environment} (b) shows the locations of grids and GWs. Figure \ref{fig:rss-heatmap} illustrates the GW IDs and the RSS distribution heatmaps for the 20 GWs. The signal coverage range for all GWs reached over 50 m. Thus, the RSS measurements at all the grid points have data from over four GWs. Meanwhile, the RSS measurement with all GWs vary over space. These facts ensure the feasibility of using RSS measurements for localization.

In the test, approximately 2,000 RSS samples from each GW were collected at each grid point. Such gridded data were further used to generate dynamic localization data through random sampling. 10,000 dynamic trajectories, each had a length of 300 steps were generated. Accordingly, there were 3,000,000 actions in the generated training data. To generate each trajectory, a grid was randomly selected as the initial point. Then, the agent started to move one grid by randomly selecting one of the nine actions in Subsection \ref{sec-mdp}. When the agent arrived a grid, a set of RSS were selected randomly through the 2,000 RSS samples from each GW and used as the RSS measurement at this step. Furthermore, to mitigate the effect of device diversity and orientation diversity, the RSS from GW 8 was selected as the datum to compute differential RSS \cite{LiY-SensJ-2019}. Meanwhile, the orientation-compensation model in \cite{LiY-SensJ-2019} was used to correct the RSS measurements.

\subsection{DRL Training}

The generated localization data were used to train the DRL. The related parameters were listed in Table \ref{tab:dqn-parameter}.

\begin{table}
\centering
\begin{tabular}{p{5.5 cm} p{1.5 cm} }
\hline
\textbf{Parameter} & \textbf{Value}   \\ \hline
\textbf{Scenario Parameter}       \\ 
Number of grid columns  & 28      \\ 
Number of grid rows  & 16      \\ 
Size of grids  & 5 m by 5 m       \\ 
Number of actions  &  9      \\ 
Number of landmark points (i.e., GWs)   & 20      \\ \hline
\textbf{Algorithm Parameter}       \\ 
Replay memory size $N_{ep}$  & 10000      \\ 
Replay start size $N_{st}$   & 2500      \\ 
Minibatch size $N_{mb}$    & 200      \\ 
Number of sample per target network update $G$  & 100      \\ 
Discount factor $\gamma$  &  0.9       \\ 
DNN learning rate  &  0.001       \\ 
Initial exploration rate $\epsilon_i$    &  1.0       \\ 
Final exploration rate $\epsilon_f$  & 0.05      \\  \hline
\end{tabular}
\caption{Values of parameters in DRL}
\label{tab:dqn-parameter}
\end{table}

\begin{figure}
\centering
\includegraphics[width=0.40 \textwidth]{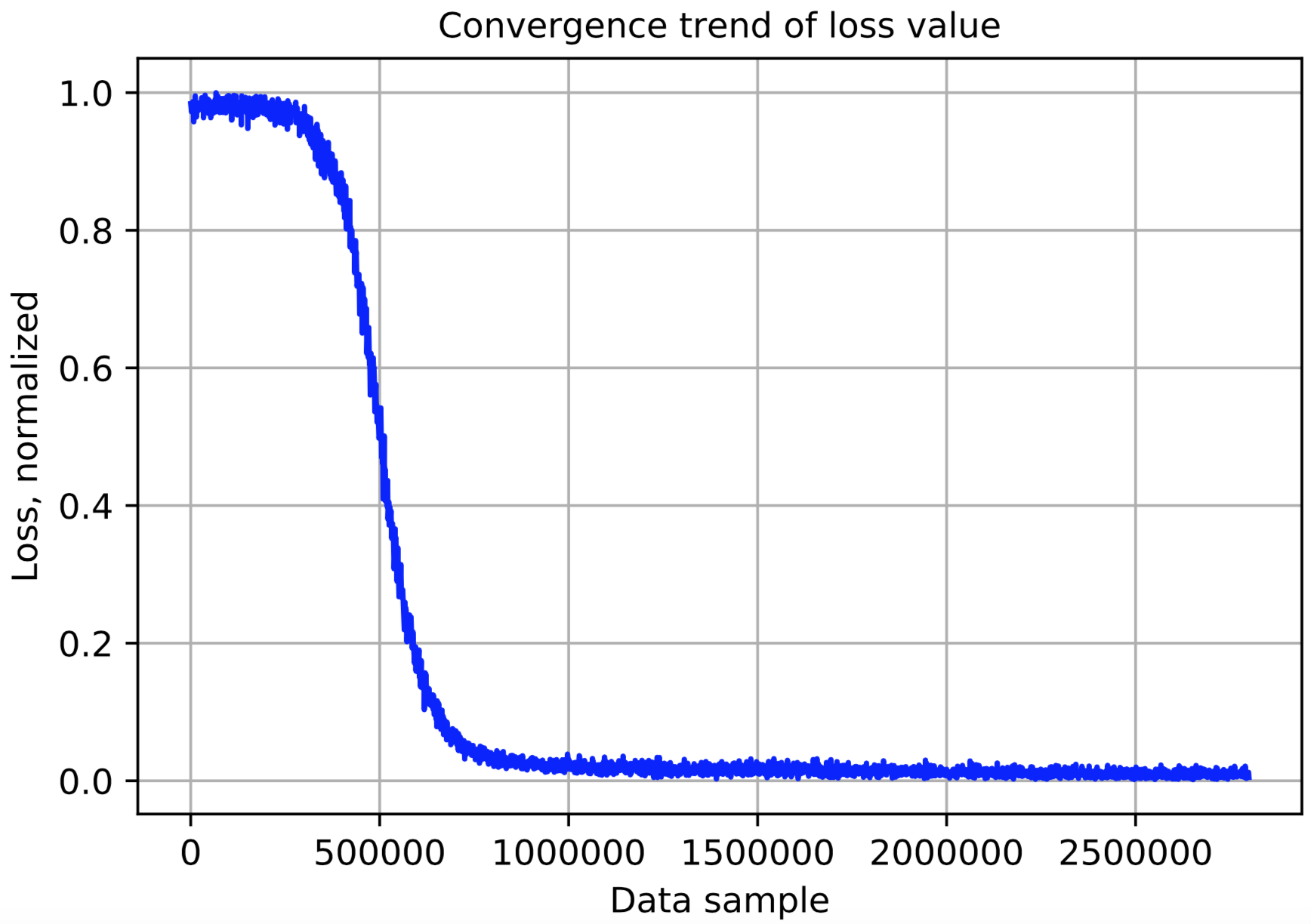}
\caption{Convergence trend of loss value over training time period}
\label{fig:loss}
\end{figure}

The data processing environment was Python 3.6 with the TensorFlow library \cite{TensorFlow}. An DNN with two hidden layers, each had 200 neurons, were used in the DQN. By running on a Macbook Pro that had a processer of 2.5 GHz Intel Core i7 and memory of 16 GB 1600 MHz DDR3, around 55 hours were taken to complete the training. Figure \ref{fig:loss} demonstrates the normalized loss function value over the training time period. It is indicated that the convergence needed around 1,000,000 samples. 

\subsection{DRL Localization}

\begin{figure}
\centering
\includegraphics[width=0.45 \textwidth]{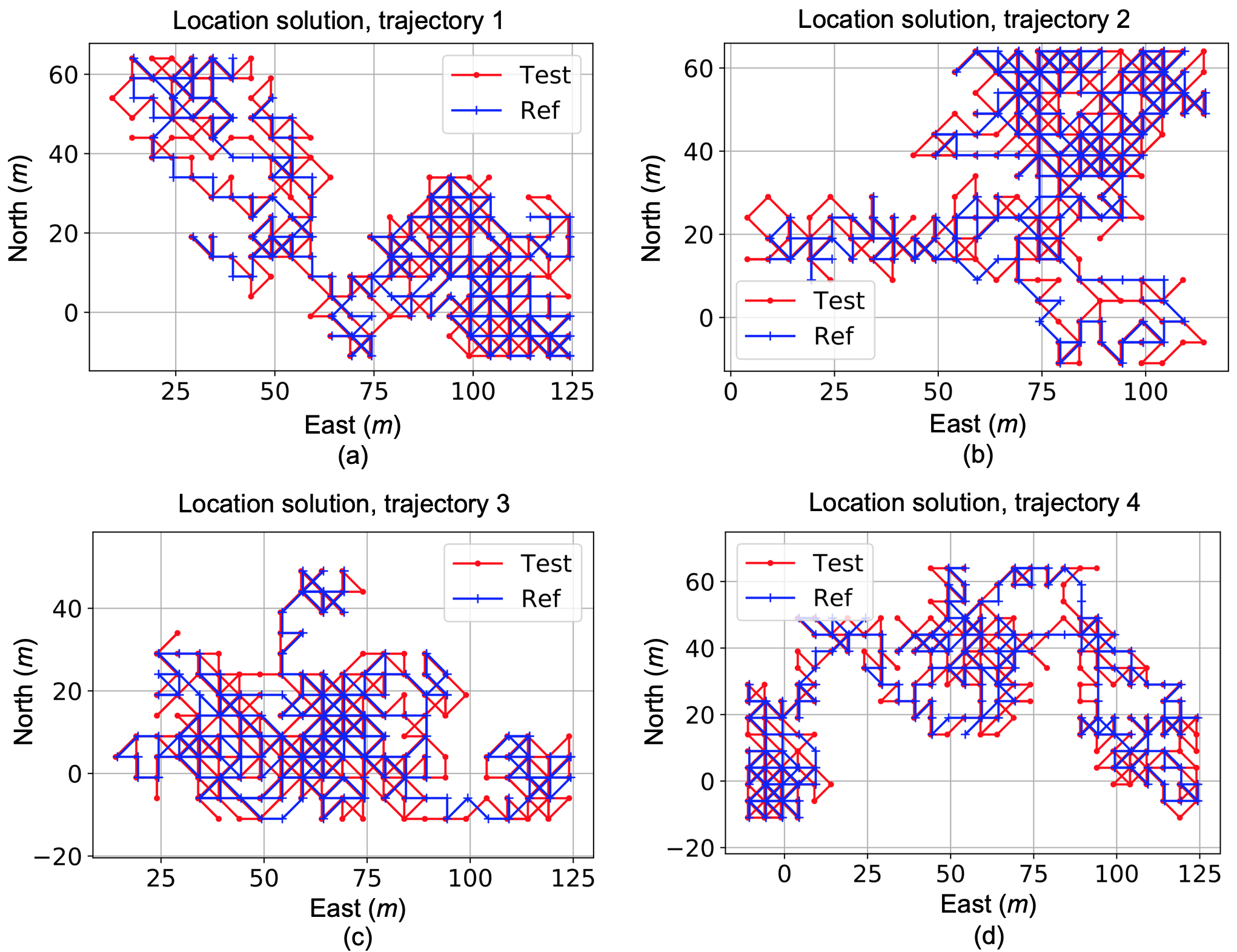}
\caption{Example of DRL-based wireless localization solutions}
\label{fig:loc-sol-traj}
\end{figure}

The trained model was used for localization. The method in Subsection \ref{sec-test-des} was used to generated 100 test trajectories, each had a length of 300 steps. Figure \ref{fig:loc-sol-traj} illustrates the localization solutions of four example trajectories. On a relatively large spatial scale (e.g., the 100 m level spatial scale), the localization solution had a similar trend with the reference trajectories. This outcome indicates the potential of using DRL for wireless localization in the long term. On the other hand, on the 10 m level spatial scale, the action output from the DRL may deviate from the actual agent movement. This phenomenon may be caused by factors such as RSS fluctuations.

\begin{figure}
\centering
\includegraphics[width=0.43 \textwidth]{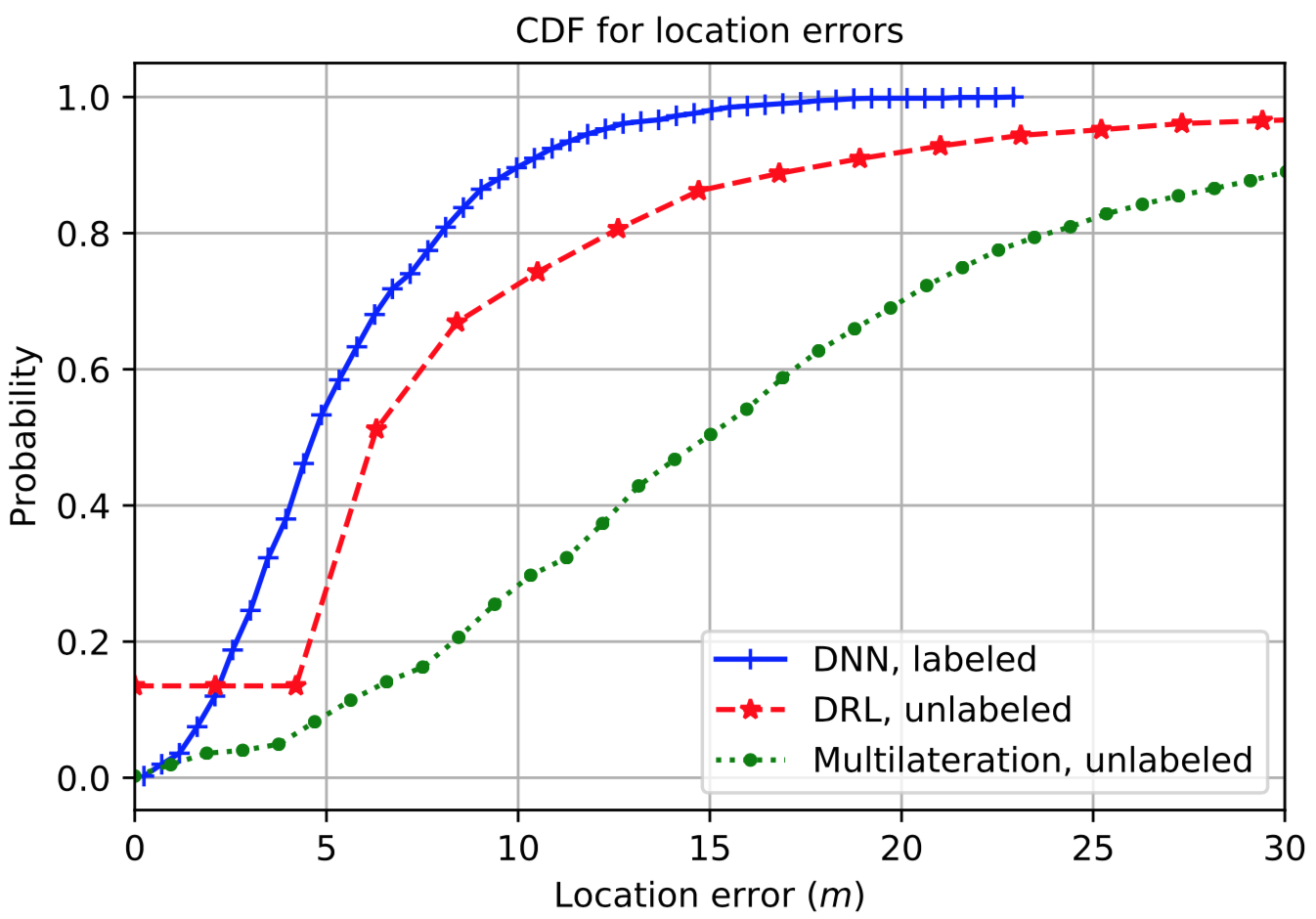}
\caption{CDF curves of location errors }
\label{fig:cdf-loc-err}
\end{figure}

\begin{figure}
\centering
\includegraphics[width=0.42 \textwidth]{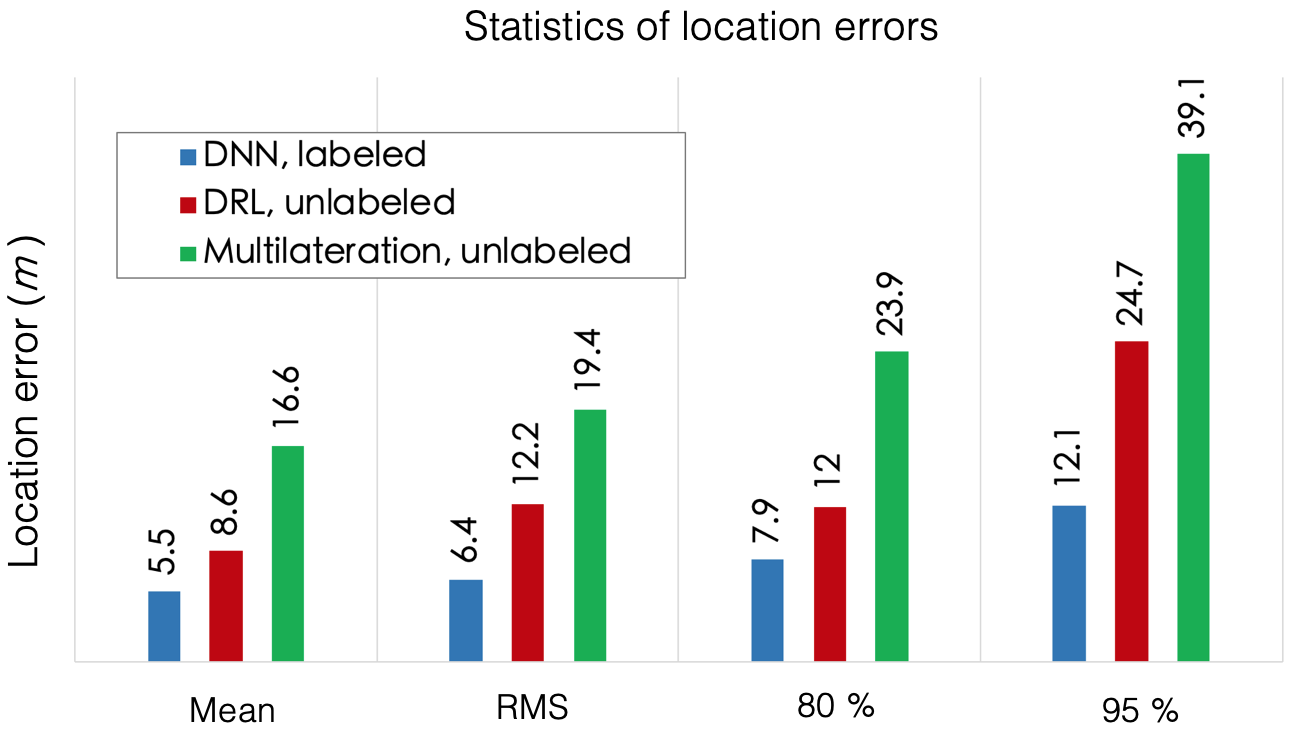}
\caption{Statistics of location errors}
\label{fig:statistics-loc-err}
\end{figure}

For comparison, the localization solutions from two comparison methods were computed. One method was DNN \cite{Zhang-Liu-2016} that uses supervised data and the other was multilateration \cite{ZhuangY-sens-2016} with unsupervised data. The former method provided a reference for the achievable localization accuracy with the test data, while the later indicated the localization accuracy when unsupervised data was used. Both comparison approaches used training and testing data that are same to the DRL-based method. On the other hand, only the supervised DNN method used the known location labels in the database-training step. The first comparison method was implemented by using a DNN with two hidden layers, each had 200 neurons. The second comparison method was applied by setting the path-loss model parameters for all GWs at experience values ($n$=2, $b$=-50). Figure \ref{fig:cdf-loc-err} demonstrates the cumulative distribution function (CDF) curves of localization errors from 100 test trajectories. Figure \ref{fig:statistics-loc-err} shows the location errors statistics, including the mean, root mean square (RMS), and the 80 \% and 95 \% quantile values. 

Figures \ref{fig:cdf-loc-err} and \ref{fig:statistics-loc-err} indicate that 

\begin{itemize}
\item The location errors from the DRL-based method had an RMS and 95 \% quantile values of 12.2 m and 24.7 m, respectively. These values were respectively 59.0 \% and 36.8 \% smaller than those from the unsupervised multilateration method (RMS 19.4 m and 95 \% in 39.1 m). This outcome indicates a positive effect by using the DRL-based method to train a DQN by using unlabeled data, and using the obtained model for localization.

\item On the other hand, the RMS and 95 \% quantile values of the DRL-based localization errors were respectively 90.6 \% and 104.1 \% higher than those from the supervised DNN method (RMS 6.4 m and 95 \% in 12.1 m). Such result indicates that the localization performance of the unsupervised DRL-based method was still significantly lower than that of the supervised DNN method. The DRL-based localization method may be further enhanced by approaches such as improving the MDP modeling (e.g., the reward-setting mechanism), improving the DRL framework, and introducing geometrical localization models.
\end{itemize}

Moreover, the following experience and insights were obtained from the tests.

\begin{itemize}
\item The DRL algorithm is data-driven and thus can be implemented without a priori motion model. An advantage for this characteristic is that such self-supervised method is suitable for complex environments that are difficult to modeling and setting parameters. On the other hand, such data-driven methods require a large amount of data and a heavy computational load (e.g., tens of hours in training for even a small scenario). To accelerate computation, the use of DRL-based localization may need support from future AI hardware and chips. Meanwhile, the DRL method is highly dependent on the quality of data. Although the DRL method itself has a well-developed exploration mechanism that may mitigate the issue of over-training, this issue is difficult to eliminate. One method for further alleviating this issue is to integrate with geometrical localization approaches and motion models. 

\item The data in this research was randomly sampled from in-field IoT data. Thus, the used data was closer to real-world situations when compared to the simulated data in the majority of existing works on DRL-based navigation and localization. However, the data in this research still cannot fully reflect the performance of the algorithms in real-world IoT localization scenarios. One main reason is that real IoT localization data may be degraded by more environmental (e.g., multipath and obstruction), motion (e.g., motion diversity), and data (e.g., data loss, database outage) factors. A future work will be using real IoT big data for evaluating AI-based localization methods.

\item The DRL algorithm itself is being enhanced due to its research and use in numerous fields. However, similar to many other AI algorithms, an DRL module is similar to a black box for most users. It is difficult to understand and adjust the internal algorithms explicitly. This factor is a potential obstacle to the study of DRL-based localization.
\end{itemize}

\section{Conclusions}

This paper presents an unsupervised wireless localization method by using the DRL framework. Through processing field-testing data from 48 BT5 smart ear tags in a pasture, which had a size of 120 m by 70 m and 20 GWs, the proposed method provided location errors that had RMS and 95 \% quantile values of 12.2 m and 24.7 m, which were respectively 59.0 \% and 36.8 \% lower than those by using an unsupervised multilateration method. Such outcome indicates a positive effect and the potential for using the DRL-based method for wireless localization. On the other hand, the RMS and 95 \% quantile values of the location errors from the proposed method were respectively 90.6 \% and 104.1 \% higher than those from the supervised DNN method. This phenomenon indicates the possibility and necessity to improve the DRL-based localization algorithm in the future. 

Meanwhile, the experimental verification process reflected several pros and cons of using DRL for localization. Its advantages include the capability to involve previous localization data and long-term rewards, the possibility to implement localization without geometrical modeling and parameterization of the environment, and the convenience of using the most state-of-the-art DRL platforms and algorithms. The challenges include the dependency on a large amount of data, the heavy computational load, and the black-box issue. The DRL-based localization method may be further enhanced by approaches such as improving the MDP modeling (e.g., the reward-setting mechanism), improving the DRL framework, and introducing geometrical localization models.

\section{Acknowledgements}

The authors would like to thank Dr. Zhe He and Dr. Yuqi Li for designing the IoT system and devices, and Daming Zhang and Stan Chan for conducting tests and data pre-processing.






\end{document}